\pdfoutput=1

\documentclass[11pt,table,xcdraw,usenames]{article}
\usepackage[preprint]{acl}
\usepackage{enumitem}
\usepackage{times}
\usepackage{latexsym}
\usepackage{amsmath}
\usepackage[T1]{fontenc}

\usepackage[utf8]{inputenc}
\usepackage{microtype}

\usepackage{inconsolata}

\usepackage{graphicx}
\usepackage{lipsum}
\usepackage{multirow} 
\usepackage{colortbl}
\usepackage{longtable}
\usepackage{arydshln}
\usepackage{tabularx}
\usepackage{booktabs}
\usepackage{caption}
\usepackage{amsmath}
\usepackage{amssymb}
\usepackage{dsfont}
\usepackage{subfigure}
\usepackage{xspace}

\usepackage{xcolor}
\usepackage{ulem}

\title{ReviewRL: Towards Automated Scientific Review with RL}

\author{
 \textbf{Sihang Zeng\textsuperscript{2}}\thanks{Equal contribution.}, 
 \textbf{Kai Tian\textsuperscript{1}}\footnotemark[1], 
 \textbf{Kaiyan Zhang\textsuperscript{1}}\footnotemark[1], 
 \textbf{Yuru Wang\textsuperscript{1}}
 \\
 \textbf{Junqi Gao\textsuperscript{3}}, 
 \textbf{Runze Liu\textsuperscript{1,3}}, 
 \textbf{Sa Yang\textsuperscript{4}}, 
 \textbf{Jingxuan Li\textsuperscript{5}}
 \\
 \textbf{Xinwei Long\textsuperscript{1}}, 
 \textbf{Jiaheng Ma\textsuperscript{6}}, 
 \textbf{Biqing Qi\textsuperscript{3}}\thanks{Corresponding author.}, 
 \textbf{Bowen Zhou\textsuperscript{1,3}}\footnotemark[2]
 \\
 \textsuperscript{1}Tsinghua University,
 \textsuperscript{2}University of Washington,
 \textsuperscript{3}Shanghai AI Laboratory
 \\
 \textsuperscript{4}Peking University,
 \textsuperscript{5}Harbin Engineering University,
 \textsuperscript{6}Beijing Institute of Technology
 \\
 \texttt{\href{mailto:qibiqing@pjlab.org.cn}{qibiqing@pjlab.org.cn}}, \texttt{\href{mailto:zhoubowen@tsinghua.edu.cn}{zhoubowen@tsinghua.edu.cn}}
}

\begin{document}
\maketitle
\begin{abstract}
    Peer review is essential for scientific progress but faces growing challenges due to increasing submission volumes and reviewer fatigue. Existing automated review approaches struggle with factual accuracy, rating consistency, and analytical depth, often generating superficial or generic feedback lacking the insights characteristic of high-quality human reviews. We introduce ReviewRL, a reinforcement learning framework for generating comprehensive and factually grounded scientific paper reviews. Our approach combines: (1) an ArXiv-MCP retrieval-augmented context generation pipeline that incorporates relevant scientific literature, (2) supervised fine-tuning that establishes foundational reviewing capabilities, and (3) a reinforcement learning procedure with a composite reward function that jointly enhances review quality and rating accuracy. Experiments on ICLR 2025 papers demonstrate that ReviewRL significantly outperforms existing methods across both rule-based metrics and model-based quality assessments. 
    ReviewRL establishes a foundational framework for RL-driven automatic critique generation in scientific discovery, demonstrating promising potential for future development in this domain. The implementation of ReviewRL will be released at \href{https://github.com/TsinghuaC3I/MARTI/tree/main/examples/reviewrl}{GitHub}.
\end{abstract}

\section{Introduction}

Peer review is critical for scientific progress, ensuring that published research meets rigorous standards of quality, validity, and significance. However, the growing volume of submissions to academic conferences and journals has created unsustainable pressure on the review system, leading to reviewer fatigue, inconsistent evaluations, and increasingly long review cycles \citep{hosseini2023fighting,kim2025position}. For instance, top-tier AI conferences like NeurIPS and ICLR now process thousands of submissions annually, requiring tens of thousands of reviews \citep{kim2025position}. This explosion in scientific output has amplified the need for automated tools to assist or augment the peer review process.

\begin{figure}[!t]
    \centering
    \includegraphics[width=\linewidth]{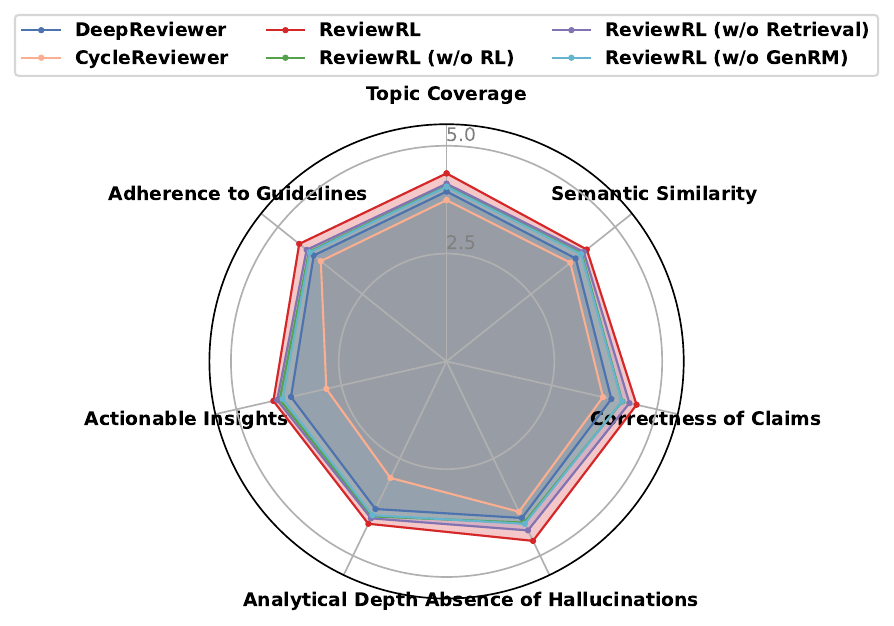}
    \caption{Evaluation results of ReviewRL under the criteria of ReviewEval~\citep{kirtani2025revieweval}}
    \label{fig:model-based-eval}
\end{figure}

Recent advances in large language models (LLMs) have created promising opportunities for AI-assisted scientific assessment. These models can analyze complex scientific texts, identify methodological strengths and weaknesses, and generate structured feedback at scale \citep{weng2024cycleresearcher,lu2024ai,zhu2025deepreview,qi2024largelanguagemodelsbiomedical}. However, existing approaches to automated paper reviewing face three significant challenges. First, they often struggle to maintain factual accuracy and provide evidence-based critiques that connect the paper to relevant prior work \citep{zhou2024llm}. Second, they tend to overestimate paper quality, assigning ratings that are inconsistently aligned with human judgment \citep{yu2025your}. Third, they frequently generate superficial or generic reviews lacking the analytical depth and actionable insights characteristic of human reviews \citep{shin2025automatically}.

Recent research has demonstrated the effectiveness of reinforcement learning (RL) in enhancing LLMs' reasoning capabilities. Models like DeepSeek-R1 \citep{guo2025deepseek} have achieved impressive performance improvements through carefully designed RL training regimes, while innovations such as Group Relative Policy Optimization \citep{shao2024deepseekmath} and Reinforce++ \citep{hu2025reinforce++} have made RL more efficient and stable for LLM training. Concurrently, the Model Context Protocol (MCP) has emerged as a standardized communication framework that enables LLMs to interact seamlessly with external knowledge sources \citep{hou2025model}, facilitating accurate information retrieval and reducing hallucinations. The combination of enhanced reasoning through RL and factual grounding via MCP-based retrieval offers a promising approach to addressing the limitations of current automated review systems.

In this paper, we introduce ReviewRL, a reinforcement learning framework for generating comprehensive, factually grounded, and constructively critical scientific paper reviews. Our approach combines three key components: (1) a ArXiv-MCP based retrieval-augmented context generation pipeline that identifies and incorporates relevant scientific literature to support factual assessments, (2) a supervised fine-tuning (SFT) phase that establishes foundational reviewing capabilities and initial rating alignment, and (3) a RL optimization procedure that jointly enhances review quality and rating accuracy. Through this integrated approach, ReviewRL produces reviews that not only emulate human-like analytical depth but also provide ratings that consistently align with human judgments.
Our experiments demonstrate that ReviewRL significantly outperforms existing approaches across both rule-based metrics and model-based assessments of review quality. We further examine the importance of each component through comprehensive ablation studies, revealing that both retrieval augmentation and our composite reward formulation contribute substantially to ReviewRL's superior performance. To our knowledge, this represents the first successful application of reinforcement learning to enhance both the quality and rating consistency of automated scientific peer reviews. Our contributions include as follows:

1) We introduce ReviewRL, a novel framework that integrates RL for automatic paper review generation. ReviewRL comprises three key components: ArxivMCP, context-aware fine-tuning, and composed reward RL training.

2) Unlike previous approaches such as DeepSeek-R1, which rely on rule-based rewards, we find such rewards insufficient for review generation, where structural coherence and content quality are paramount. To address this, we design a comprehensive reward system incorporating both rule-based metrics and judge-model-based evaluations, effectively mitigating the limitations of purely rule-based methods.

3) Compared to prior work, ReviewRL achieves superior performance in context-awareness, factual consistency, and review depth. This framework represents a preliminary step toward RL-driven automatic critique generation in scientific discovery.

\begin{figure*}
    \centering
    \includegraphics[width=\linewidth]{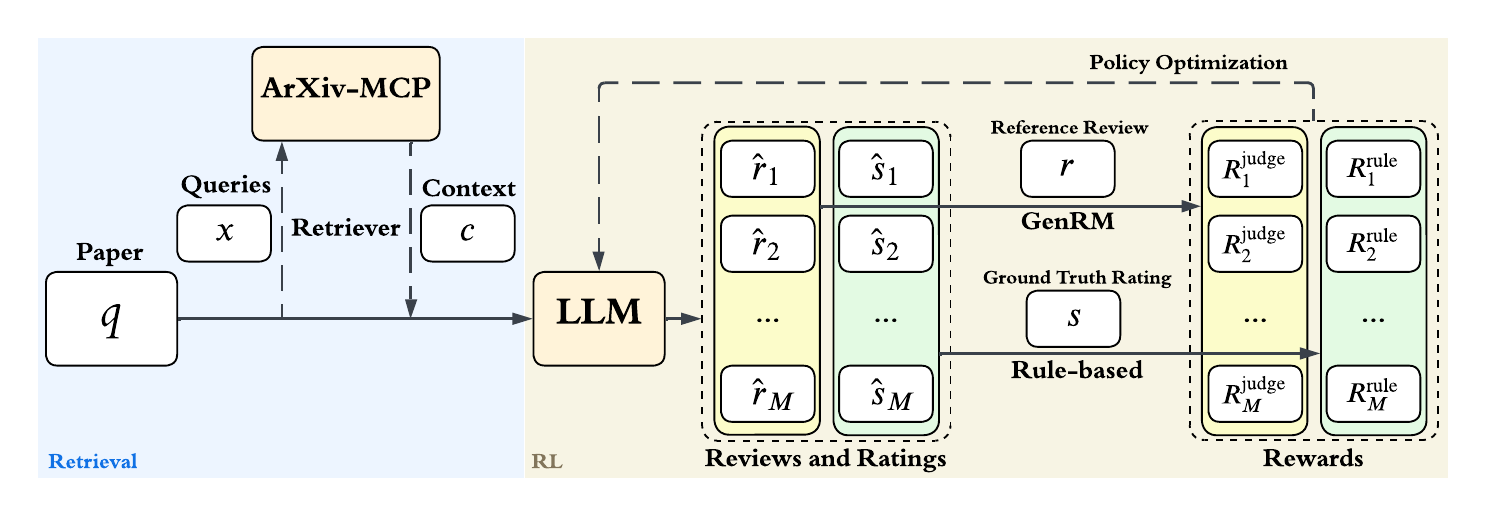}
    \caption{Overview of ReviewRL, including Arxiv-MCP, SFT warm up, and RL optimization.}
    \label{fig:main_fig}
\end{figure*}

\section{Related Works}

\paragraph{LLM for Paper Review} Recent advancements have explored the use of LLMs to automate and enhance the academic peer review process. Early efforts, such as PeerRead \citep{kang2018dataset} and NLPeer \citep{dycke2022nlpeer}, provided foundational datasets and benchmarks for review generation and analysis. Building upon these resources, systems like Reviewer2 \citep{gao2024reviewer2} proposed a two-stage framework involving aspect prompt generation and review generation to improve the specificity and coverage of generated reviews. CycleResearcher \citep{weng2024cycleresearcher} and The AI Scientist \citep{lu2024ai} have introduced end-to-end frameworks that simulate the entire research lifecycle, including manuscript drafting and iterative peer review, where their reviewer modules are trained via supervised fine-tuning or operate through agentic inference. More recently, DeepReviewer \citep{zhu2025deepreview} is trained through SFT using long chain-of-thought (CoT) data to enhance its reasoning ability. Despite these advancements, challenges remain in ensuring the factualness, reasoning depth and rating consistency of LLM-generated reviews.

\paragraph{Reinforcement Learning for LLMs} 
Reinforcement Learning (RL)~\citep{sutton1998reinforcement} plays a pivotal role in enhancing the instruction-following capabilities of Large Language Models (LLMs), particularly through approaches like Reinforcement Learning from Human Feedback (RLHF)~\citep{ouyang2022training}. RLHF aligns foundation models with human preferences, typically leveraging algorithms such as Proximal Policy Optimization (PPO)~\citep{schulman2017proximal} or Direct Preference Optimization~\citep{rafailov2023direct}, where precise preference modeling is essential.
Recent advancements have demonstrated RL's effectiveness in improving reasoning abilities in Large Reasoning Models (LRMs), such as DeepSeek-R1~\citep{guo2025deepseek}, through rule-based reward mechanisms, as exemplified by GRPO~\citep{shao2024deepseekmath}. Unlike RLHF, which is generally applied to open-domain instructions, GRPO is specifically designed to promote extended Chain-of-Thought (CoT)~\citep{wei2022chain} reasoning, particularly in mathematical problem-solving scenarios.
Benefiting from its simplicity and effectiveness, GRPO has been successfully applied across diverse domains, including vision understanding and generation~\citep{liu2025visual,team2025kimi,xue2025dancegrpo}, agentic search and planning~\citep{li2025webthinker,jin2025search,zhang2025agent}, and beyond.
However, the potential of RL methods like GRPO to enhance review and critique generation~\citep{whitehouse2025j1} still need more exploration.

\section{Methodology}

\subsection{Task Formulation}
\label{sec:task}

Given a target paper \( q \), the automated scientific review task is defined as generating a comprehensive review \( r \), including the paper’s strengths and weaknesses, and a rating \( s \). To ensure high-quality review generation, we formulate ReviewRL's workflow as a retrieval-augmented generation (RAG) \citep{lewis2020retrieval} and a LLM reasoning process. This process mimics the cognitive steps of human reviewers—retrieving relevant domain knowledge, analyzing the paper in context, and making evaluative judgments. Specifically, a retriever model \( R \) generates three queries $x$ and identifies a set of relevant contextual papers \( c \) through searching, formulated as \( q \xrightarrow{R} x, c \). An LLM-based reviewer \( \pi \) then reasons over the paper and the retrieved context to produce an intermediate thinking process \( z \), represented as \( (q, c) \xrightarrow{\pi} z \). Finally, the LLM generates the review and rating based on the paper, the retrieved context, and the reasoning trace, i.e., \( (q, c, z) \xrightarrow{\pi} (r, s) \).

In the following sections, we present the components of ReviewRL as shown in Figure~\ref{fig:main_fig}. We first introduce the RAG pipeline (Section~\ref{sec:paper_retrieval}) that accurately identifies contextually relevant literature given the target paper. We then describe our training strategy for ReviewRL, which combines SFT (Section~\ref{sec:sft}) and RL (Section~\ref{sec:rl}) to enhance reasoning capabilities.

\subsection{Context Retrieval}
\label{sec:paper_retrieval}

For each paper $q$, we retrieve relevant contextual information $c$ using a retrieval pipeline $R$. Following a two-step approach inspired by the novelty verification in DeepReviewer \citep{zhu2025deepreview}, we first generate query questions $x$ and then retrieve relevant contexts $c$ from ArXiv. This method leverages an LLM agentic workflow and integrates the Model Context Protocol (MCP) for efficient context retrieval and generation.

Specifically, we employ Qwen3-8B \citep{yang2025qwen3} to analyze the target paper $q$ and generate three query questions $x$ that probe the paper's novelty, methodology, and relationship to prior work. These queries are formulated as natural language questions rather than keywords, allowing for more nuanced retrieval of relevant literature. For example, a query might ask ``What recent papers have proposed reinforcement learning for LLM-based paper reviews?'' rather than simply searching for ``reinforcement learning LLM reviews.''

We implement the retrieval functionality through ArXiv-MCP\footnote{\url{https://github.com/blazickjp/arxiv-mcp-server}}, an open-source Model Context Protocol server that provides LLMs with standardized access to the arXiv repository. ArXiv-MCP enables efficient paper search, filtering, and full-text retrieval without requiring low-level API implementation. The server processes the generated queries and returns structured information including paper metadata, abstracts, and relevant full-text excerpts.

The retrieval execution is orchestrated by Qwen-Agent\footnote{\url{https://github.com/QwenLM/Qwen-Agent}}, an agent framework built on the Qwen model family that provides function calling and tool orchestration capabilities. Qwen-Agent sequentially routes each query to ArXiv-MCP and manages the retrieval results. The retrieved contexts $c$ undergo post-processing to remove tool invocation artifacts and are consolidated into a coherent format that includes: (1) query-response pairs, (2) bibliographic information of retrieved papers, and (3) relevance-ranked excerpts from these papers. This processed context is then concatenated with the original paper representation to form the input for the review generation model.

\subsection{Supervised Finetuning}
\label{sec:sft}
Given a paper \( q \) and its retrieved context \( c \), the next step is to generate the review \( r \) and the corresponding rating \( s \) using a policy model $\pi$. While our goal is to enhance this process using RL, direct RL application on base models presents challenges. These models typically overestimate paper quality compared to human reviewers \citep{yu2025your}, leading to uninformative trajectories with weak reward signals and unstable RL training. Empirically, we observe that without proper initialization, RL training exhibits a cold-start problem characterized by training collapse and performance degradation (e.g., generated ratings collapse around $6$).

To mitigate this, we first apply SFT on long CoT data to initialize the RL policy with essential review-writing capabilities. This strategy is inspired by similar practice in DeepSeek-R1 \citep{guo2025deepseek} and Kimi-k1.5 \citep{team2025kimi}. We leverage data derived from DeepReview-13k \citep{zhu2025deepreview}, a high-quality dataset comprising long CoT reviews and accurate rating annotations, as the cold-start data for training the model. Specifically, we use the ICLR 2024 portion of the dataset and preprocess it to fit our task definition. We include their novelty verification results and the corresponding queries from the best mode in the input, and use the final meta review as the output, with intermediate analysis regarded as the long CoT thinking process. We train for 2 epochs on top of the model Qwen2.5-7B-Instruct \citep{qwen2.5}. Different from previous work without RL, in our framework, SFT serves two primary goals: (1) to equip the policy model with foundational reasoning ability to perform structured and reasoned peer reviews, and (2) to align predicted scores with human ratings, thereby stabilizing downstream RL training and preventing early-stage collapse.

\subsection{Reinforcement Learning}
\label{sec:rl}

Following the SFT phase, we conduct large-scale reinforcement learning (RL) to further enhance the reasoning capabilities of the LLM reviewer. Paper reviewing is a non-verifiable problem with a partially verifiable outcome—the numerical rating—where both the \textit{review quality} and \textit{rating consistency} with human judgments are essential. Prior work has demonstrated the effectiveness of rule-based outcome rewards in improving LLM reasoning \cite{guo2025deepseek}. However, our experiments show that relying solely on a rating consistency reward leads to overly generic reviews lacking analytical depth and actionable insights, indicating insufficient reasoning ability.

To jointly optimize the review quality and rating consistency, we design a composite reward that integrates rule-based rewards with a generative reward model (GenRM) \citep{zhang2024generative}, which prioritizes reviews with high rating consistency, format adherence, and strong analytical depth.

\paragraph{Rule-Based Rewards}  
We define two rule-based reward components: rating consistency reward and format reward. The rating consistency reward \( R_{rc} \) is computed using a Gaussian kernel:

\begin{align}
    R_{rc} = \exp\left(-\frac{(s - \hat{s})^2}{2\sigma^2}\right)
\end{align}

where \( s \) denotes the ground-truth rating, obtained by averaging human-assigned scores for the given paper, and \( \hat{s} \) is the rating predicted by the model.
The format adherence reward \( R_f \) penalizes outputs that omit essential structural components. Let \( \mathcal{S} \) be the set of required elements, including a reasoning block (delimited by \texttt{<think>} and \texttt{</think>}), summary, strengths, and weaknesses:

\begin{align}
    R_f = - \sum_{s \in \mathcal{S}} \mathds{1} (s \text{ is missing})
\end{align}

The final rule-based reward is given by:

\begin{align}
    R^{\text{rule}} = \text{clip}\left(\alpha \cdot R_{rc} + \beta \cdot R_f, 0, 1\right)
\end{align}

where \( \alpha \) and \( \beta \) are hyperparameters that balance the importance of rating consistency and format completeness. This reward formulation encourages the generation of outputs that are both aligned with human ratings and structurally well-formed.

\paragraph{GenRM-based Rewards}  
Following prior work \citep{seed2025seed,hogan2024debate}, we employ a GenRM \( \pi_{\text{judge}} \) to evaluate the quality of the LLM-generated review $\hat{r}$ against a reference $r$. The reward is derived from the win rate, based on the agreement that LLM-as-a-judge can reliably assess relative response quality \citep{zheng2023judgingllmasajudgemtbenchchatbot}. In our framework, \( \pi_{\text{judge}} \) evaluates reviews across multiple dimensions: factual accuracy, completeness, level of detail, comparison with related work, constructiveness, and clarity. The GenRM reward \( R^{\text{judge}} \) is defined as:
\begin{align}
R^{\text{judge}} = \begin{cases}
1 & \text{if $\hat{r}$ is preferred} \\
0 & \text{if $r$ is preferred}
\end{cases}
\end{align}

The final reward signal is computed as a weighted combination of the rule-based reward and the GenRM reward:
\begin{align}
R^{\text{final}} = \gamma R^{\text{rule}} + (1 - \gamma) R^{\text{judge}}
\end{align}

\paragraph{RL Training Data}  
We construct the RL training dataset using papers from top-tier machine learning conferences, such as ICLR and ACL, sourced from the raw data of Reviewer2 \citep{gao2024reviewer2} and the best mode split of DeepReview-13k \citep{zhu2025deepreview}. For each paper \( q \), we retrieve the context \( c \) using the method described in Section~\ref{sec:paper_retrieval}. Ratings from different conferences are normalized to a common scale of 1–10, and the ground truth rating \( s \) is computed as the average of scores from multiple human reviewers.
Reference reviews \( r \) are derived as follows: for each paper from Reviewer2, we summarize multiple human reviews using DeepSeek-R1-Distill-Qwen-32B into a single formatted review; for DeepReview-13k, we use the meta-reviews from the best mode split. ICLR 2025 data are excluded to avoid data leakage. Dataset statistics are reported in Table~\ref{tab:data_statistics}. Because a large proportion of ground truth ratings fall between 5 and 6, we apply a \textit{balancing} preprocessing step that downsamples papers with mid-range ratings (5–6) and upsamples those with more extreme ratings. This strategy emphasizes papers with highly positive or negative assessments, which tend to be more informative for learning, and helps prevent the RL model from collapsing to generic, non-discriminative ratings around the middle range.

\begin{table}[ht]
  \centering
  \resizebox{\columnwidth}{!}{%
    \begin{tabular}{lrrrrrr}
      \hline
      & \textbf{ICLR} & \textbf{NeurIPS} & \textbf{ARR} & \textbf{COLING} & \textbf{CONLL} & \textbf{ACL}\\
      \textbf{Year} & 2017-2024 & 2021-2022 & 2022 & 2020 & 2016 & 2017 \\
      \hline
      \textbf{Count} & 13312 & 3994 & 336 & 82 & 22 & 131 \\
      \textbf{Avg. \#Token} &9854&10275&9153&8138&7888&8571 \\
      \hline
    \end{tabular}%
  }
  \caption{RL training data statistics}
  \label{tab:data_statistics}
\end{table}

Therefore, the RL training data comprises tuples of \( (q, c, s, r) \), without access to the intermediate reasoning steps that lead to the review \( r \) and rating \( s \). This setup encourages the policy model to explore its own reasoning trajectories that produce high-quality reviews and ratings consistent with human judgments.

\paragraph{RL Training Setting}  
The policy model \( \pi \) is initialized from the supervised finetuned model \( \pi_{\text{sft}} \) to ensure stable learning and prevent cold-start collapse. We adopt the Reinforce++ algorithm \citep{hu2025reinforce++}. $\pi_{\text{judge}}$ is a Qwen2.5-14B-Instruct model. Training details are shown in Appendix \ref{app:training}.

\section{Experiments}

\subsection{Evaluation Data}  
We construct the evaluation set by sampling 472 papers from the ICLR~2025 review corpus. For each paper, the ground truth rating is computed as the average of scores assigned by all human reviewers. To ensure fair evaluation across the full rating spectrum, we sample papers such that the distribution of average ratings is approximately uniform across the rating scale.

\subsection{Evaluation Metrics}
We employ two families of quantitative metrics:
\textbf{(i) rule-based} metrics, and
\textbf{(ii) model-based} metrics.

\subsubsection{Rule-based Quantitative Metrics}
We evaluate the model using both rating-level and pairwise-level metrics. For score prediction, we compute the mean squared error (MSE) and Spearman rank correlation between predicted scores and ground truth ratings. For pairwise paper evaluation, following JudgeLRM~\citep{chen2025judgelrm}, we assess the model's ability to rank paper quality using three pairwise metrics: \emph{relation}, \emph{absolute}, and \emph{confidence}. These respectively measure directional consistency with human rankings, score closeness to ground truth, and discriminative strength in differentiating papers of varying quality. Concordance index is also reported as a global ranking metric. Formal definitions of the pairwise metrics are provided in Appendix~\ref{sec:pairwise-metrics}.

\subsubsection{Model-based Quantitative Metrics}
While rule-based metrics focus on the accuracy of the generated ratings, it is equally important to assess whether the generated reviews emulate human-written reviews and provide constructive, content-rich feedback. To this end, we adopt an LLM-as-a-judge framework \citep{zheng2023judgingllmasajudgemtbenchchatbot} inspired by the ReviewEval benchmark \citep{kirtani2025revieweval}, evaluating review quality across seven dimensions. Each dimension is rated on a 1-5 scale and aims to capture a distinct aspect of human-aligned peer reviewing:

\begin{itemize}[leftmargin=*,labelindent=0pt,itemsep=2pt,topsep=3pt]
    \item \textbf{Topic Coverage:} Does the AI-generated review comprehensively address the main topics and arguments of the paper? Does it cover aspects typically emphasized by human reviewers?
    
    \item \textbf{Semantic Similarity:} Does the review capture the core critique and suggestions of a plausible human review, even if phrased differently?
    
    \item \textbf{Correctness of Claims:} Are the statements in the review factually accurate with respect to the paper’s content? Does the review avoid misinterpretations or incorrect representations of the methodology, results, or conclusions?
    
    \item \textbf{Absence of Hallucinations:} Does the review refrain from introducing information or claims not supported by the paper?
    
    \item \textbf{Analytical Depth:} Does the review demonstrate deep engagement with the research? This includes evaluating methodological rigor, identifying logical gaps, interpreting results, and contextualizing contributions within related work.
    
    \item \textbf{Actionable Insights:} Does the review provide specific, constructive suggestions for improving the paper? Are the recommendations practical and clearly articulated?
    
    \item \textbf{Adherence to Guidelines:} Does the review follow standard academic review criteria such as originality, significance, methodological soundness, clarity, and ethical compliance (if applicable)?
\end{itemize}

This evaluation framework enables a comprehensive assessment of the model’s ability to perform nuanced and human-aligned paper reviewing beyond surface-level metrics. Llama-3.3-70B-Instruct is used as the judge model.

\subsection{Baselines}

We compare against three classes of baselines. \textbf{Open-source \textit{instruction}} models (e.g., Qwen-2.5-Instruct) and \textbf{Open-source \textit{reasoning}} models (e.g., Qwen\,3) provides baseline paper review performance with basic instruction following ability and enhanced reasoning capabilities. Additionally, \textbf{SFT} models trained on public peer-review datasets, such as CycleReviewer-8B, are included to highlight the performance gain achieved by our RL-enhanced model over purely supervised approaches.

\section{Results}

\begin{table*}[]

\caption{Rule-based evaluation results. ReviewRL consistently outperforms baseline methods.}
\label{tab:rule-based-results}
\resizebox{\textwidth}{!}{
\begin{tabular}{lllllll}
\toprule
\textbf{Model}                         & \textbf{MSE $\downarrow$}     & \textbf{Spearman $\uparrow$} & \textbf{Pair-Relation $\uparrow$} & \textbf{Pair-Absolute $\uparrow$} & \textbf{Pair-Confidence $\uparrow$} & \textbf{Concordance $\uparrow$}  \\ \midrule
\multicolumn{7}{c}{\cellcolor[HTML]{CBCEFB}\textit{Open Source Instruct}}                                                        \\ \midrule
Qwen2.5-7B-Instruct           & 12.024 & 0.158   & 0.514       & 0.051        & 0.138        & 0.668 \\
Qwen2.5-32B-Instruct          & 9.847  & 0.147   & 0.538       & 0.055        & 0.345        & 0.575 \\
Qwen2.5-72B-Instruct          & 9.418  & 0.325   & 0.529       & 0.074        & 0.318        & 0.705 \\
Llama-3.3-70B-Instruct        & 9.839  & 0.285   & 0.539       & 0.061        & 0.286        & 0.687 \\ \midrule
\multicolumn{7}{c}{\cellcolor[HTML]{FFCE93}\textit{Open Source Reasoning}}                                                        \\ \midrule
DeepSeek-R1-Distill-Qwen-7B   & 9.247  & 0.062   & 0.512       & 0.063        & 0.399        & 0.527 \\
DeepSeek-R1-Distill-Qwen-14B  & 10.064 & 0.271   & 0.525       & 0.065        & 0.281        & 0.683 \\
DeepSeek-R1-Distill-Qwen-32B  & 6.463  & 0.341   & 0.569       & 0.097        & 0.414        & 0.677 \\
DeepSeek-R1-Distill-Llama-70B & 9.021  & 0.389   & 0.539       & 0.065        & 0.336        & 0.747 \\
QwQ-32B                       & 5.440  & 0.425   & 0.585       & 0.128        & 0.402        & 0.743 \\
Qwen3-8B                      & 4.852  & 0.237   & 0.567       & 0.157        & 0.294        & 0.649 \\
Qwen3-14B                     & 8.348  & 0.371   & 0.534       & 0.072        & 0.391       & 0.706 \\ 
Qwen3-32B                     & 9.613  & 0.415   & 0.528       & 0.182        & \textbf{0.462}        & 0.753 \\\midrule
\multicolumn{7}{c}{\cellcolor[HTML]{FFFFC7}\textit{SFT Training}}                                                                \\ \midrule
CycleReviewer-ML-Llama3.1-8B & 4.409  & 0.482   & 0.495       & 0.192        & 0.355        & 0.743 \\ 
DeepReviewer-7B              & 3.445  & 0.539   & \textbf{0.639}       & 0.245        & 0.245        & 0.710 \\ 
\textbf{ReviewRL-7B (w/o RL)}              & 2.829  & 0.335   & 0.528       & 0.135        & 0.260        & 0.644 \\ \hline
\multicolumn{7}{c}{\cellcolor[HTML]{ECF4FF}\textit{RL Training}}                                                                          \\ \midrule
\textbf{ReviewRL-7B}                     & \textbf{2.585}  & \textbf{0.634}   & 0.579       & \textbf{0.249}        & 0.360       & \textbf{0.806} \\
\bottomrule

\end{tabular}
}

\end{table*}

\subsection{Rule-based Evaluation Results}
Table~\ref{tab:rule-based-results} presents rule-based evaluation results. Open-source \textit{instruct} models perform weakest overall, showing poor rating accuracy and limited ranking capability, even at larger scales. Open-source \textit{reasoning} models improve upon pairwise metrics, particularly in discriminative strength (Pair-Confidence), but still lag in MSE. SFT models trained on peer review datasets demonstrate significant gains in the alignment with human ratings—e.g., DeepReviewer-7B achieves the best Pair-Relation score. Our proposed ReviewRL model achieves the strongest performance across the board. The performance gap between ReviewRL and its SFT-only counterpart highlights the effectiveness of reinforcement learning in optimizing rating consistency.

\subsection{Model-based Evaluation Results}

Figure~\ref{fig:model-based-eval} shows the model-based evaluation results across seven review quality dimensions. ReviewRL consistently outperforms all baselines, particularly in dimensions that require deeper reasoning and reliability, such as analytical depth. Compared to its supervised-only counterpart, ReviewRL exhibits clear improvements in all dimensions, highlighting the benefits of RL in refining review generation. These results further demonstrate that RL leads to more informative, faithful, and constructive reviews.

\subsection{RL Training Dynamics}
\begin{figure*}[ht]
    \centering
    \includegraphics[width=\linewidth]{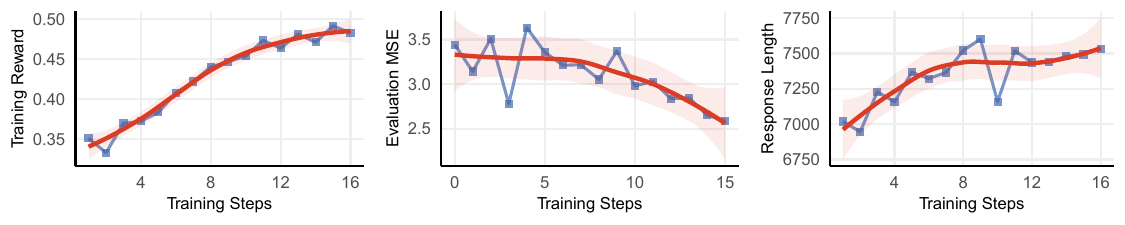}
    \caption{Training Dynamics of RL}
    \label{fig:training_dynamics}
\end{figure*}

\begin{figure}[ht]
    \centering
    \includegraphics[width=\linewidth]{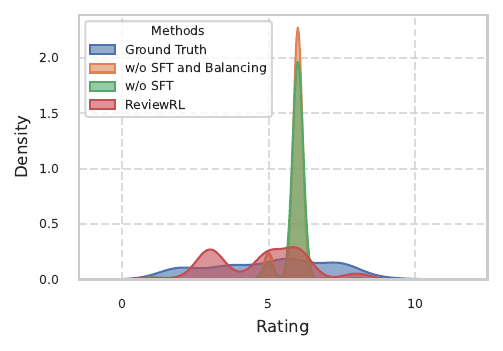}
    \caption{Rating Distributions}
    \label{fig:cold_start}
\end{figure}

We analyze the RL training dynamics of ReviewRL to provide insights for future training recipes in LLM reviewer models.
\paragraph{Training Curves} 
Figure \ref{fig:training_dynamics} illustrates the training dynamics of ReviewRL across three key metrics. As training progresses, the training reward steadily increases, indicating that the policy is effectively optimizing for the reward function. Simultaneously, the response length grows in the earlier stages and stabilizes after approximately step 10, suggesting that the model learns to generate more detailed outputs. The evaluation MSE decreases consistently over training steps, confirming that the learned policy generalizes better to held-out data and produces more accurate review ratings. These trends collectively demonstrate the effectiveness of the RL training procedure.
\paragraph{Cold-Start Phase}
We observe a cold-start issue in ReviewRL when RL training is initiated without proper policy initialization. As shown in Figure~\ref{fig:cold_start}, training from scratch leads to rating collapse, where the model predominantly outputs generic scores around 6 and fails to differentiate between input papers. Our data balancing strategy for RL training, which upweights examples with extreme ground-truth ratings, partially mitigates this issue but remains insufficient alone. In contrast, combining SFT with data balancing enables ReviewRL to produce ratings across the full spectrum, exhibiting stronger discrimination and alignment with ground-truth annotations.
\paragraph{Reward Shaping}
We conduct an ablation study where only the rule-based reward is used during RL training. As shown in Figure~\ref{fig:model-based-eval}, removing the GenRM reward leads to no significant improvement over the SFT baseline across model-based evaluation metrics, with slightly lower actionable insights. This highlights the critical role of GenRM in guiding the policy model to generate high-quality reviews with sufficient details and reliable reasoning. The result aligns with recent findings that emphasize the importance of GenRM or judge models for learning in RL settings involving non-verifiable tasks.

\subsection{Context Retrieval}
We evaluate the context retrieval module from two perspectives: the quality of the retrieved context $c$ and its impact on review generation.

\paragraph{Quality of Retrieved Context}
For each generated query $x$, we compare two responses: an ArXiv-MCP retrieval-augmented answer $c$ and a vanilla answer $c_0$ generated without external search. Three independent LLM judges assess each pair based on three criteria: (1) \textit{Factual Accuracy}—correctness and alignment with real-world facts, (2) \textit{Evidence Quality}—sufficiency and relevance of supporting evidence, and (3) \textit{Clarity \& Coherence}—readability, organization, and logical flow. We report the win rate where the retrieval-augmented answer is better than the vanilla answer.

\begin{figure}[!t]
    \centering
    \includegraphics[width=\linewidth]{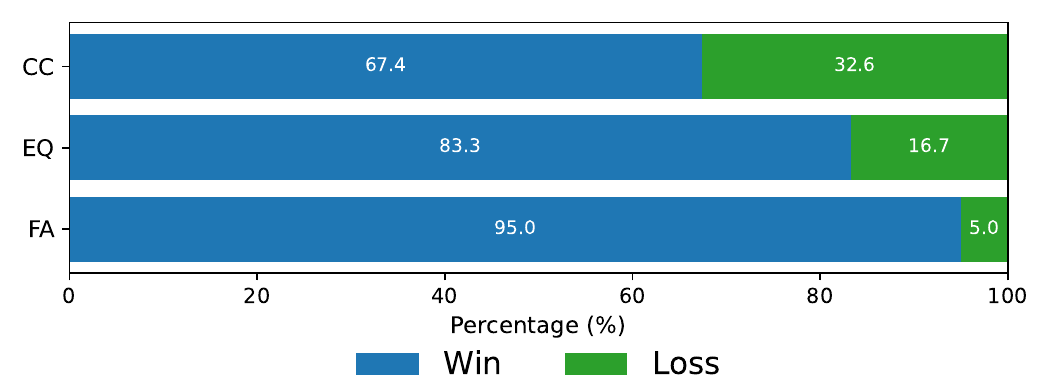}
    \caption{Win/Loss percentages of the retrieval answer (Win = blue) versus the non‑retrieval answer (Loss = green) across three evaluation dimensions.}
    \label{fig:eval_bar}
\end{figure}

As shown in Figure~\ref{fig:eval_bar}, retrieval-augmented responses outperform vanilla responses across all criteria. In terms of factual accuracy, 95.0\% of comparisons favor the retrieval variant, indicating a substantial reduction in hallucinations. Evidence quality shows an 83.3\% win rate, suggesting effective integration of retrieved citations. While the gain in clarity and coherence is smaller (67.4\%), retrieval-augmented responses are still preferred in the majority of cases, implying that additional evidence does not hinder readability. Overall, retrieval consistently enhances context quality, with the most pronounced effect on factual accuracy.

\paragraph{Impact on Review Generation}
We conduct inference on ReviewRL under the setting where no retrieved context is provided as input (ReviewRL w/o Retrieval). As shown in Figure \ref{fig:model-based-eval}, without retrieval, we observe performance degradation across all metrics, especially for the factualness metrics including correctness of claims and absence of hallucinations. The analytical depth and topic coverage also shrinks, potentially because comparison between the paper and related works may not be effectively conducted without retrieval.

\section{Conclusion}
In this paper, we introduced ReviewRL, a reinforcement learning framework for automating scientific paper reviews. Our approach integrates context retrieval, supervised fine-tuning, and reinforcement learning to generate high-quality, human-aligned paper reviews with accurate ratings. Experimental results on ICLR 2025 papers demonstrate that ReviewRL significantly outperforms existing methods across both rule-based and model-based evaluation metrics. We established a principled methodology for combining SFT and RL in non-verifiable reasoning tasks, showing that properly initialized policy models can effectively learn from composite rewards without experiencing cold-start issues. Additionally, we demonstrated the critical role of retrieved context in enhancing review factuality and analytical depth, substantiating the effectiveness of our ArXiv-MCP retrieval pipeline.

\section*{Limitations}
Our framework relies on the accessibility and comprehensiveness of ArXiv as the primary knowledge source, which may provide insufficient context for papers exploring emerging research directions or highly specialized domains with limited representation in the repository. Additionally, although our composite reward function effectively balances rating consistency and review quality, it remains challenging to fully capture the nuanced aspects of human peer review that extend beyond our seven evaluation dimensions. Domain-specific criteria and conference-specific review expectations, which often involve implicit knowledge and norms within academic communities, may not be adequately represented in our current reward formulation, potentially limiting ReviewRL's adaptation to specialized venues or interdisciplinary research areas.

\section*{Ethical Considerations}
To ensure the ethical development and use of the ReviewRL system, a multifaceted approach has been implemented. During training, we have carefully curated training data and designed the system's reward function to prioritize factual accuracy, analytical depth, and rating consistency, thereby reducing unintended biases and risks. Crucially, ReviewRL is intended to support, not replace, human reviewers, with its outputs serving as drafts for expert evaluation and refinement. For transparency, we will open-source the system, accompanied by detailed documentation on its architecture and training. We will require the users to disclose their affiliation and intended use, fostering accountability and a feedback mechanism for continuous improvement. Furthermore, the system's context retrieval component is designed to maximize coverage and minimize citation bias, and its evaluation metrics consider diverse aspects of review quality beyond simple accuracy, aiming to harness ReviewRL's benefits while proactively mitigating potential harms in the peer review process.

\bibliography{main}

\begin{thebibliography}{37}
\providecommand{\natexlab}[1]{#1}

\bibitem[{Chen et~al.(2025)Chen, Hu, Zou, Wu, Wang, Hooi, and
  He}]{chen2025judgelrm}
Nuo Chen, Zhiyuan Hu, Qingyun Zou, Jiaying Wu, Qian Wang, Bryan Hooi, and
  Bingsheng He. 2025.
\newblock Judgelrm: Large reasoning models as a judge.
\newblock \emph{arXiv preprint arXiv:2504.00050}.

\bibitem[{Dycke et~al.(2022)Dycke, Kuznetsov, and Gurevych}]{dycke2022nlpeer}
Nils Dycke, Ilia Kuznetsov, and Iryna Gurevych. 2022.
\newblock Nlpeer: A unified resource for the computational study of peer
  review.
\newblock \emph{arXiv preprint arXiv:2211.06651}.

\bibitem[{Gao et~al.(2024)Gao, Brantley, and Joachims}]{gao2024reviewer2}
Zhaolin Gao, Kiant{\'e} Brantley, and Thorsten Joachims. 2024.
\newblock Reviewer2: Optimizing review generation through prompt generation.
\newblock \emph{arXiv preprint arXiv:2402.10886}.

\bibitem[{Guo et~al.(2025)Guo, Yang, Zhang, Song, Zhang, Xu, Zhu, Ma, Wang, Bi
  et~al.}]{guo2025deepseek}
Daya Guo, Dejian Yang, Haowei Zhang, Junxiao Song, Ruoyu Zhang, Runxin Xu,
  Qihao Zhu, Shirong Ma, Peiyi Wang, Xiao Bi, and 1 others. 2025.
\newblock Deepseek-r1: Incentivizing reasoning capability in llms via
  reinforcement learning.
\newblock \emph{arXiv preprint arXiv:2501.12948}.

\bibitem[{Hogan(2024)}]{hogan2024debate}
Brendan Hogan. 2024.
\newblock \href {https://github.com/brendanhogan/debate-framework} {Debate
  framework for language model training}.

\bibitem[{Hosseini and Horbach(2023)}]{hosseini2023fighting}
Mohammad Hosseini and Serge~PJM Horbach. 2023.
\newblock Fighting reviewer fatigue or amplifying bias? considerations and
  recommendations for use of chatgpt and other large language models in
  scholarly peer review.
\newblock \emph{Research integrity and peer review}, 8(1):4.

\bibitem[{Hou et~al.(2025)Hou, Zhao, Wang, and Wang}]{hou2025model}
Xinyi Hou, Yanjie Zhao, Shenao Wang, and Haoyu Wang. 2025.
\newblock Model context protocol (mcp): Landscape, security threats, and future
  research directions.
\newblock \emph{arXiv preprint arXiv:2503.23278}.

\bibitem[{Hu(2025)}]{hu2025reinforce++}
Jian Hu. 2025.
\newblock Reinforce++: A simple and efficient approach for aligning large
  language models.
\newblock \emph{arXiv preprint arXiv:2501.03262}.

\bibitem[{Jin et~al.(2025)Jin, Zeng, Yue, Yoon, Arik, Wang, Zamani, and
  Han}]{jin2025search}
Bowen Jin, Hansi Zeng, Zhenrui Yue, Jinsung Yoon, Sercan Arik, Dong Wang, Hamed
  Zamani, and Jiawei Han. 2025.
\newblock Search-r1: Training llms to reason and leverage search engines with
  reinforcement learning.
\newblock \emph{arXiv preprint arXiv:2503.09516}.

\bibitem[{Kang et~al.(2018)Kang, Ammar, Dalvi, Van~Zuylen, Kohlmeier, Hovy, and
  Schwartz}]{kang2018dataset}
Dongyeop Kang, Waleed Ammar, Bhavana Dalvi, Madeleine Van~Zuylen, Sebastian
  Kohlmeier, Eduard Hovy, and Roy Schwartz. 2018.
\newblock A dataset of peer reviews (peerread): Collection, insights and nlp
  applications.
\newblock \emph{arXiv preprint arXiv:1804.09635}.

\bibitem[{Kim et~al.(2025)Kim, Lee, and Lee}]{kim2025position}
Jaeho Kim, Yunseok Lee, and Seulki Lee. 2025.
\newblock Position: The ai conference peer review crisis demands author
  feedback and reviewer rewards.
\newblock \emph{arXiv preprint arXiv:2505.04966}.

\bibitem[{Kirtani et~al.(2025)Kirtani, Garg, Prasad, Singhal, Mandal, and
  Kumar}]{kirtani2025revieweval}
Chhavi Kirtani, Madhav~Krishan Garg, Tejash Prasad, Tanmay Singhal, Murari
  Mandal, and Dhruv Kumar. 2025.
\newblock Revieweval: An evaluation framework for ai-generated reviews.
\newblock \emph{arXiv preprint arXiv:2502.11736}.

\bibitem[{Lewis et~al.(2020)Lewis, Perez, Piktus, Petroni, Karpukhin, Goyal,
  K{\"u}ttler, Lewis, Yih, Rockt{\"a}schel et~al.}]{lewis2020retrieval}
Patrick Lewis, Ethan Perez, Aleksandra Piktus, Fabio Petroni, Vladimir
  Karpukhin, Naman Goyal, Heinrich K{\"u}ttler, Mike Lewis, Wen-tau Yih, Tim
  Rockt{\"a}schel, and 1 others. 2020.
\newblock Retrieval-augmented generation for knowledge-intensive nlp tasks.
\newblock \emph{Advances in neural information processing systems},
  33:9459--9474.

\bibitem[{Li et~al.(2025)Li, Jin, Dong, Qian, Zhu, Wu, Wen, and
  Dou}]{li2025webthinker}
Xiaoxi Li, Jiajie Jin, Guanting Dong, Hongjin Qian, Yutao Zhu, Yongkang Wu,
  Ji-Rong Wen, and Zhicheng Dou. 2025.
\newblock Webthinker: Empowering large reasoning models with deep research
  capability.
\newblock \emph{arXiv preprint arXiv:2504.21776}.

\bibitem[{Liu et~al.(2025)Liu, Sun, Zang, Dong, Cao, Duan, Lin, and
  Wang}]{liu2025visual}
Ziyu Liu, Zeyi Sun, Yuhang Zang, Xiaoyi Dong, Yuhang Cao, Haodong Duan, Dahua
  Lin, and Jiaqi Wang. 2025.
\newblock Visual-rft: Visual reinforcement fine-tuning.
\newblock \emph{arXiv preprint arXiv:2503.01785}.

\bibitem[{Lu et~al.(2024)Lu, Lu, Lange, Foerster, Clune, and Ha}]{lu2024ai}
Chris Lu, Cong Lu, Robert~Tjarko Lange, Jakob Foerster, Jeff Clune, and David
  Ha. 2024.
\newblock The ai scientist: Towards fully automated open-ended scientific
  discovery.
\newblock \emph{arXiv preprint arXiv:2408.06292}.

\bibitem[{Ouyang et~al.(2022)Ouyang, Wu, Jiang, Almeida, Wainwright, Mishkin,
  Zhang, Agarwal, Slama, Ray et~al.}]{ouyang2022training}
Long Ouyang, Jeffrey Wu, Xu~Jiang, Diogo Almeida, Carroll Wainwright, Pamela
  Mishkin, Chong Zhang, Sandhini Agarwal, Katarina Slama, Alex Ray, and 1
  others. 2022.
\newblock Training language models to follow instructions with human feedback.
\newblock \emph{Advances in neural information processing systems},
  35:27730--27744.

\bibitem[{Qi et~al.(2024)Qi, Zhang, Tian, Li, Chen, Zeng, Hua, Jinfang, and
  Zhou}]{qi2024largelanguagemodelsbiomedical}
Biqing Qi, Kaiyan Zhang, Kai Tian, Haoxiang Li, Zhang-Ren Chen, Sihang Zeng,
  Ermo Hua, Hu~Jinfang, and Bowen Zhou. 2024.
\newblock \href {https://arxiv.org/abs/2407.08940} {Large language models as
  biomedical hypothesis generators: A comprehensive evaluation}.
\newblock \emph{Preprint}, arXiv:2407.08940.

\bibitem[{Rafailov et~al.(2023)Rafailov, Sharma, Mitchell, Manning, Ermon, and
  Finn}]{rafailov2023direct}
Rafael Rafailov, Archit Sharma, Eric Mitchell, Christopher~D Manning, Stefano
  Ermon, and Chelsea Finn. 2023.
\newblock Direct preference optimization: Your language model is secretly a
  reward model.
\newblock \emph{Advances in Neural Information Processing Systems},
  36:53728--53741.

\bibitem[{Schulman et~al.(2017)Schulman, Wolski, Dhariwal, Radford, and
  Klimov}]{schulman2017proximal}
John Schulman, Filip Wolski, Prafulla Dhariwal, Alec Radford, and Oleg Klimov.
  2017.
\newblock Proximal policy optimization algorithms.
\newblock \emph{arXiv preprint arXiv:1707.06347}.

\bibitem[{Seed et~al.(2025)Seed, Yuan, Yue, Wang, Zuo, Chen, Yan, Xu, Zhang,
  Liu et~al.}]{seed2025seed}
ByteDance Seed, Yufeng Yuan, Yu~Yue, Mingxuan Wang, Xiaochen Zuo, Jiaze Chen,
  Lin Yan, Wenyuan Xu, Chi Zhang, Xin Liu, and 1 others. 2025.
\newblock Seed-thinking-v1. 5: Advancing superb reasoning models with
  reinforcement learning.
\newblock \emph{arXiv preprint arXiv:2504.13914}.

\bibitem[{Shao et~al.(2024)Shao, Wang, Zhu, Xu, Song, Bi, Zhang, Zhang, Li, Wu
  et~al.}]{shao2024deepseekmath}
Zhihong Shao, Peiyi Wang, Qihao Zhu, Runxin Xu, Junxiao Song, Xiao Bi, Haowei
  Zhang, Mingchuan Zhang, YK~Li, Y~Wu, and 1 others. 2024.
\newblock Deepseekmath: Pushing the limits of mathematical reasoning in open
  language models.
\newblock \emph{arXiv preprint arXiv:2402.03300}.

\bibitem[{Shin et~al.(2025)Shin, Tang, Lee, Kim, Lim, Cho, Hong, Lee, and
  Kim}]{shin2025automatically}
Hyungyu Shin, Jingyu Tang, Yoonjoo Lee, Nayoung Kim, Hyunseung Lim, Ji~Yong
  Cho, Hwajung Hong, Moontae Lee, and Juho Kim. 2025.
\newblock Automatically evaluating the paper reviewing capability of large
  language models.
\newblock \emph{arXiv preprint arXiv:2502.17086}.

\bibitem[{Sutton et~al.(1998)Sutton, Barto et~al.}]{sutton1998reinforcement}
Richard~S Sutton, Andrew~G Barto, and 1 others. 1998.
\newblock \emph{Reinforcement learning: An introduction}, volume~1.
\newblock MIT press Cambridge.

\bibitem[{Team et~al.(2025)Team, Du, Gao, Xing, Jiang, Chen, Li, Xiao, Du, Liao
  et~al.}]{team2025kimi}
Kimi Team, Angang Du, Bofei Gao, Bowei Xing, Changjiu Jiang, Cheng Chen, Cheng
  Li, Chenjun Xiao, Chenzhuang Du, Chonghua Liao, and 1 others. 2025.
\newblock Kimi k1. 5: Scaling reinforcement learning with llms.
\newblock \emph{arXiv preprint arXiv:2501.12599}.

\bibitem[{Team(2024)}]{qwen2.5}
Qwen Team. 2024.
\newblock \href {https://qwenlm.github.io/blog/qwen2.5/} {Qwen2.5: A party of
  foundation models}.

\bibitem[{Wei et~al.(2022)Wei, Wang, Schuurmans, Bosma, Xia, Chi, Le, Zhou
  et~al.}]{wei2022chain}
Jason Wei, Xuezhi Wang, Dale Schuurmans, Maarten Bosma, Fei Xia, Ed~Chi, Quoc~V
  Le, Denny Zhou, and 1 others. 2022.
\newblock Chain-of-thought prompting elicits reasoning in large language
  models.
\newblock \emph{Advances in neural information processing systems},
  35:24824--24837.

\bibitem[{Weng et~al.(2024)Weng, Zhu, Bao, Zhang, Wang, Zhang, and
  Yang}]{weng2024cycleresearcher}
Yixuan Weng, Minjun Zhu, Guangsheng Bao, Hongbo Zhang, Jindong Wang, Yue Zhang,
  and Linyi Yang. 2024.
\newblock Cycleresearcher: Improving automated research via automated review.
\newblock \emph{arXiv preprint arXiv:2411.00816}.

\bibitem[{Whitehouse et~al.(2025)Whitehouse, Wang, Yu, Li, Weston, Kulikov, and
  Saha}]{whitehouse2025j1}
Chenxi Whitehouse, Tianlu Wang, Ping Yu, Xian Li, Jason Weston, Ilia Kulikov,
  and Swarnadeep Saha. 2025.
\newblock J1: Incentivizing thinking in llm-as-a-judge via reinforcement
  learning.
\newblock \emph{arXiv preprint arXiv:2505.10320}.

\bibitem[{Xue et~al.(2025)Xue, Wu, Gao, Kong, Zhu, Chen, Liu, Liu, Guo, Huang
  et~al.}]{xue2025dancegrpo}
Zeyue Xue, Jie Wu, Yu~Gao, Fangyuan Kong, Lingting Zhu, Mengzhao Chen, Zhiheng
  Liu, Wei Liu, Qiushan Guo, Weilin Huang, and 1 others. 2025.
\newblock Dancegrpo: Unleashing grpo on visual generation.
\newblock \emph{arXiv preprint arXiv:2505.07818}.

\bibitem[{Yang et~al.(2025)Yang, Li, Yang, Zhang, Hui, Zheng, Yu, Gao, Huang,
  Lv et~al.}]{yang2025qwen3}
An~Yang, Anfeng Li, Baosong Yang, Beichen Zhang, Binyuan Hui, Bo~Zheng, Bowen
  Yu, Chang Gao, Chengen Huang, Chenxu Lv, and 1 others. 2025.
\newblock Qwen3 technical report.
\newblock \emph{arXiv preprint arXiv:2505.09388}.

\bibitem[{Yu et~al.(2025)Yu, Luo, Madusu, Lal, and Howard}]{yu2025your}
Sungduk Yu, Man Luo, Avinash Madusu, Vasudev Lal, and Phillip Howard. 2025.
\newblock Is your paper being reviewed by an llm? a new benchmark dataset and
  approach for detecting ai text in peer review.
\newblock \emph{arXiv preprint arXiv:2502.19614}.

\bibitem[{Zhang et~al.(2024)Zhang, Hosseini, Bansal, Kazemi, Kumar, and
  Agarwal}]{zhang2024generative}
Lunjun Zhang, Arian Hosseini, Hritik Bansal, Mehran Kazemi, Aviral Kumar, and
  Rishabh Agarwal. 2024.
\newblock Generative verifiers: Reward modeling as next-token prediction.
\newblock \emph{arXiv preprint arXiv:2408.15240}.

\bibitem[{Zhang et~al.(2025)Zhang, Yang, Shu, Wen, and Sang}]{zhang2025agent}
Yuxiang Zhang, Yuqi Yang, Jiangming Shu, Xinyan Wen, and Jitao Sang. 2025.
\newblock Agent models: Internalizing chain-of-action generation into reasoning
  models.
\newblock \emph{arXiv preprint arXiv:2503.06580}.

\bibitem[{Zheng et~al.(2023)Zheng, Chiang, Sheng, Zhuang, Wu, Zhuang, Lin, Li,
  Li, Xing, Zhang, Gonzalez, and
  Stoica}]{zheng2023judgingllmasajudgemtbenchchatbot}
Lianmin Zheng, Wei-Lin Chiang, Ying Sheng, Siyuan Zhuang, Zhanghao Wu, Yonghao
  Zhuang, Zi~Lin, Zhuohan Li, Dacheng Li, Eric~P. Xing, Hao Zhang, Joseph~E.
  Gonzalez, and Ion Stoica. 2023.
\newblock \href {https://arxiv.org/abs/2306.05685} {Judging llm-as-a-judge with
  mt-bench and chatbot arena}.
\newblock \emph{Preprint}, arXiv:2306.05685.

\bibitem[{Zhou et~al.(2024)Zhou, Chen, and Yu}]{zhou2024llm}
Ruiyang Zhou, Lu~Chen, and Kai Yu. 2024.
\newblock Is llm a reliable reviewer? a comprehensive evaluation of llm on
  automatic paper reviewing tasks.
\newblock In \emph{Proceedings of the 2024 Joint International Conference on
  Computational Linguistics, Language Resources and Evaluation (LREC-COLING
  2024)}, pages 9340--9351.

\bibitem[{Zhu et~al.(2025)Zhu, Weng, Yang, and Zhang}]{zhu2025deepreview}
Minjun Zhu, Yixuan Weng, Linyi Yang, and Yue Zhang. 2025.
\newblock Deepreview: Improving llm-based paper review with human-like deep
  thinking process.
\newblock \emph{arXiv preprint arXiv:2503.08569}.

\end{thebibliography}

\appendix

\section{Prompts}
The prompts for both the Generation, Evaluation, and GenRM are presented in Tables \ref{tab:prompt for Gqp}, \ref{tab:Rsp}, \ref{tab:eval_model_prompt}, \ref{tab:Reep} and \ref{tab:judge_prompt}.

\section{Pairwise Metrics}
\label{sec:pairwise-metrics}

\begin{equation}
  \mathrm{P}_{\text{relation}}
    = \begin{cases}
        1.0, & \text{if } \operatorname{sgn}(s_{1}-s_{2})
                = \operatorname{sgn}(s^{\ast}_{1}-s^{\ast}_{2}), \\[4pt]
       0, & \text{otherwise}.
      \end{cases}
  \label{eq:relation}
\end{equation}

\begin{equation}
  \mathrm{P}_{\text{absolute}}
    = \begin{cases}
        1.0, & \text{if } |s_{1}-s^{\ast}_{1}| + |s_{2}-s^{\ast}_{2}| = 0, \\[4pt]
        0.6, & \begin{aligned}[t]
                 |s_{1}-s^{\ast}_{1}| + |s_{2}-s^{\ast}_{2}| \le 2,
               \end{aligned} \\[4pt]
        0,   & \text{otherwise}.
      \end{cases}
  \label{eq:absolute}
\end{equation}
\begin{equation}
  \mathrm{P}_{\text{confidence}}
    = \begin{cases}
        1.0, & \begin{aligned}[t]
                 |s_{1}-s_{2}| \ge |s^{\ast}_{1}-s^{\ast}_{2}|,
               \end{aligned} \\[4pt]
        0,   & \text{otherwise}.
      \end{cases}
  \label{eq:confidence}
\end{equation}

In this formula, \(s_1\) and \(s_2\) represent the model's output review ratings, while \(s_1^*\) and \(s_2^*\) represent the corresponding ground truth values. 
\textbf{Relation} assesses \emph{directional consistency} with human reviewers.
\textbf{Absolute} measures the \emph{score proximity} to human reviewers.
\textbf{Confidence} examines differences in \emph{discriminative strength}.

\begin{table}[htbp]
\centering
\begin{tabular}{ll}

\hline
Llama-3.1-8B-Instruct & MSE $\downarrow$  \\ \hline
SFT                   & 3.01 \\ 
Reinforce++           & 2.80 \\ \hline
\end{tabular}
\caption{MSE of \textit{Llama‑3.1‑8B‑Instruct} after SFT vs.\ Reinforce++ training (lower is better)}
\label{tab:llama-general}
\end{table}

\begin{table}[htbp]
\centering
\begin{tabular}{ll}
\hline
Qwen2.5-7B-Instruct & MSE $\downarrow$  \\ \hline
SFT                 & 2.83 \\ 
Reinforce++ (ReviewRL)         & 2.59 \\ 
PPO                 & 2.69 \\ 
GRPO                & 2.63 \\ \hline
\end{tabular}
\caption{Comparison of RL algorithms on the \textit{Qwen} backbone (lower MSE is better)}
\label{tab:rl-algo}
\end{table}

\begin{table*}[]
\centering
\resizebox{\textwidth}{!}{
\begin{tabular}{lccccccc}
\hline
{\textbf{}}                   & \multicolumn{1}{l}{\textbf{Topic Cov.}} & \multicolumn{1}{l}{\textbf{Sem.Sim.}} & \multicolumn{1}{l}{\textbf{Cor. of Claims}} & \multicolumn{1}{l}{\textbf{Abs. of Hal.}} & \multicolumn{1}{l}{\textbf{Ana. Depth}} & \multicolumn{1}{l}{\textbf{Act.Ins.}} & \multicolumn{1}{l}{\textbf{Adh. to Guide.}} \\ \hline
\textbf{DeepReviewer}             & 3.94                                    & 3.83                                  & 3.92                                        & 4.03                                      & 3.80                                     & 3.70                                   & 3.94                                        \\
\textbf{CycleReviewer}            & 3.74                                    & 3.67                                  & 3.72                                        & 3.87                                      & 3.00                                       & 2.86                                  & 3.73                                        \\
\textbf{ReviewRL}                 & \textbf{4.36}                           & \textbf{4.16}                         & \textbf{4.52}                               & \textbf{4.62}                             & \textbf{4.18}                           & \textbf{4.12}                         & \textbf{4.37}                               \\
\textbf{ReviewRL (w/o RL)}        & 4.07                                    & 4.01                                  & 4.18                                        & 4.15                                      & 3.99                                    & 3.97                                  & 4.08                                        \\
\textbf{ReviewRL (w/o Retrieval)} & 4.12                                    & 4.07                                  & 4.35                                        & 4.35                                      & 4.04                                    & 4.03                                  & 4.15                                        \\
\textbf{ReviewRL(w/o GenRM)}      & 4.05                                    & 3.99                                  & 4.17                                        & 4.18                                      & 3.96                                    & 3.90                                   & 4.06                                        \\ \hline
\end{tabular}}

\caption{Model‑based evaluation scores on seven quality dimensions for baselines, ablation variants, and our proposed \textsc{ReviewRL} system (higher is better). This table contains the same quantitative results visualised in Figure~\ref{fig:model-based-eval}.}
\label{tab:figure1_table}
\end{table*}

\begin{table*}[ht]
\centering
\small
\begin{tabularx}{\textwidth}{X}
\toprule
\underline{\textbf{\textsc{Generate queries prompt}}} \\
\midrule
You are now an academic paper review expert capable of conducting thorough analyses of research papers to provide the most reliable review results. You are now allowed to use the search tool to obtain background information on the paper—please provide three different questions. I will assist you with the search. Please present the three questions in the following format:\\
1.xxx\\
2.xxx\\
3.xxx\\
Do not include any additional content.\\
\\
Here is a research paper:\\
\{paper\} \\
\bottomrule
\end{tabularx}
\caption{Prompt for Generate queries prompt}
\label{tab:prompt for Gqp}
\end{table*}

\begin{table*}[htbp]
\centering
\small
\begin{tabularx}{\textwidth}{X}
\toprule
\underline{\textbf{\textsc{Retrieval system prompt}}} \\

\midrule
You are an academic expert who specializes in answering questions
by retrieving information from arXiv.\\

\bottomrule
\end{tabularx}
\caption{Retrieval system prompt}
\label{tab:Rsp}
\end{table*}

\begin{table*}[htbp]
\centering
\small
\begin{tabularx}{\textwidth}{X}
\toprule
\underline{\textbf{\textsc{Open source model evaluation prompt}}} \\

\midrule
Here is a research paper:\\  
\texttt{\{paper\}} \\
You are a senior reviewer for top-tier AI conferences (NeurIPS/ICML/CVPR/ACL).\\
You must be strict and professional enough.  \\
\textbf{Read the Paper Carefully:}\\
\quad Analyze each paragraph of each section critically.\\
\quad Identify any logical flaws, technical inconsistencies, missing citations, or unclear explanations.\\
\textbf{Detailed Paragraph-by-Paragraph Review:}\\
\quad Provide a detailed critique of each paragraph in every section.\\
\quad If a paragraph contains multiple issues, list them separately.\\
\quad Highlight strengths, but be critical of weaknesses.\\
\quad Use \verb|<think> </think>| tags to document your detailed thought process during the review. \\
\textbf{Comprehensive Structured Review:}\\
After the detailed paragraph-by-paragraph critique, provide a structured review using the following format:\\
\quad \#\# Summary\\
\quad (3–5 sentences: core contribution + methodology)\\
\quad \#\# Strengths\\
\quad - Bullet points focusing on: Technical merit | Novelty | Empirical validation\\
\quad \#\# Weaknesses\\
\quad - Bullet points labeled [Major] or [Minor]: Methodology flaws | Experimental issues | Presentation problems\\
\quad \#\# Rating\\
\quad One integer from: [1, 3, 5, 6, 8, 10]
(10=Strong Accept; 8=Accept; 6=Borderline Accept; 5=Borderline Reject; 3=Reject; 1=Strong Reject)\\

\bottomrule
\end{tabularx}
\caption{Open source model evaluation prompt}
\label{tab:eval_model_prompt}
\end{table*}

\begin{table*}[htbp]
\centering
\small
\begin{tabularx}{\textwidth}{X}
\toprule
\underline{\textbf{\textsc{Retrieval effectiveness evaluation prompt}}} \\

\midrule
\textbf{Factual Accuracy:}\\
\textbf{You are an extremely meticulous domain expert.}\\[0.3em]
Task: Compare Answer-A (which uses retrieval) with Answer-B (which does not) \textbf{only on factual accuracy / faithfulness}.\\[0.3em]
\textbf{Scoring rule}\\
$\bullet$ If Answer-A is fully correct or clearly more accurate than Answer-B $\rightarrow$ output 0\\
$\bullet$ If Answer-B is clearly more accurate $\rightarrow$ output 1\\
$\bullet$ If both are equally correct but Answer-A supplies extra verifiable details, still treat Answer-A as better $\rightarrow$ output 0\\[0.4em]
\textbf{Output format:} a single character 0 or 1—nothing else.\\

\midrule
\textbf{Evidence Quality:}\\
\textbf{You are an academic reviewer. Judge the two answers solely on the quality and usefulness of their evidence or citations.}\\[0.3em]
\textbf{Decision rule}\\
0 = Answer-A provides stronger or clearer evidence / citations.\\
1 = Answer-B provides stronger or clearer evidence / citations.\\
If both contain little or equivalent evidence, but Answer-A supplies extra verifiable details $\rightarrow$ output 0\\[0.4em]
Return \textbf{only} the single digit 0 or 1. Any extra text is invalid.\\

\midrule
\textbf{Clarity \& Coherence:}\\
\textbf{You are a senior instructor. Evaluate which answer demonstrates better clarity and coherence.}\\[0.3em]
\textbf{Consider}\\
$\bullet$ Is the writing easy to follow and well-organized?\\
$\bullet$ Are ideas presented in a logical order with smooth transitions?\\
$\bullet$ Is terminology explained and jargon minimized?\\
$\bullet$ Does the answer avoid unnecessary repetition or ambiguity?\\[0.4em]
If Answer-A is better clear/coherent than Answer-B $\rightarrow$ output 0; otherwise output 1.\\[0.3em]
\textbf{Output} must be exactly one character: 0 or 1.\\

\bottomrule
\end{tabularx}
\caption{Retrieval effectiveness evaluation prompt}
\label{tab:Reep}
\end{table*}

\begin{table*}[ht]
\centering
\small
\begin{tabularx}{\linewidth}{X}

\toprule
\underline{\textbf{\textsc{GenRM prompt}}} \\

\midrule
You are an expert academic peer reviewer. You will be shown the abstract/content of a research paper and two peer reviews for that paper. Your task is to determine which peer review is of higher quality based on the following criteria: \\
\vspace{0.5em}
\textbf{1. Factual Accuracy \& Soundness:} Does the review accurately understand the paper's contributions and limitations? Is the critique based on sound reasoning? \\
\textbf{2. Completeness \& Coverage:} Does the review address the core aspects of the paper (e.g., methodology, results, significance)? \\
\textbf{3. Level of Detail \& Specificity:} Does the review provide specific examples and detailed comments rather than vague statements? \\
\textbf{4. Comparison with Existing Work:} Does the review appropriately contextualize the paper within the existing literature and compare it to relevant methods? \\
\textbf{5. Constructiveness:} Is the feedback helpful for the authors to improve the paper? Is the tone professional and constructive? \\
\textbf{6. Clarity \& Organization:} Is the review well-structured and easy to understand? \\
\vspace{0.5em}
\textbf{Paper Context (Abstract/Content):} \{paper\_context\} \\
\textbf{Review 1:} \{review1\} \\
\textbf{Review 2:} \{review2\} \\
\vspace{0.5em}
Which peer review is of higher quality based on the criteria above? Respond with \textbf{EXACTLY} one of these options: \\
- REVIEW\_1\_BETTER \\
- REVIEW\_2\_BETTER \\
\vspace{0.5em}
\textbf{YOU MUST CHOOSE A BETTER REVIEW. A TIE IS NOT ALLOWED.} \\
\bottomrule
\end{tabularx}
\caption{GenRM prompt.}
\label{tab:judge_prompt}
\end{table*}

\section{Training Settings}
\label{app:training}
\subsection{RL Training}

DeepSpeed ZeRO-3 and Ray were employed for distributed reinforcement learning on dual 8$\times$A800 GPU clusters. Configuration: micro-batch size of 1, global batch size of 128, and 8-sample rollouts per prompt. Reference and actor models were colocated, with 6 GPUs allocated to the vLLM Engine and 2 GPUs to the GenRM. The composite reward used a balancing coefficient $\gamma=0.5$. Training completed in 15 optimization steps over 48 hours.
\subsection{SFT Training}
Supervised fine-tuning utilized DeepSpeed ZeRO-3 with a batch size of 8 and learning rate of 5e-6.

\subsection{Model-based Evaluation}
Table~\ref{tab:figure1_table} presents the quantitative results from Figure~\ref{fig:model-based-eval} to enable a precise comparison of system performance. ReviewRL achieves the highest score across all evaluated dimensions, with particularly significant gains in analytical depth and factuality. We observe a strong positive correlation across all dimensions, indicating that systems excelling in one metric tend to excel in others. This finding suggests that generating high-quality scientific reviews is a multifaceted task that requires a comprehensive set of integrated capabilities rather than proficiency in isolated skills.

\subsection{Comparison across RL Algorithms}
\label{subsec:rl-algo}

To explore the robustness of \textsc{ReviewRL} to alternative RL algorithms, we train the model with \textbf{PPO} and \textbf{GRPO} in addition to our default \textbf{Reinforce++}.  
As reported in Table~\ref{tab:rl-algo}, RL models consistently outperform the SFT baseline under all three algorithms, validating the robustness of our composite reward.  

\subsection{Model Generality across Architectures}
\label{subsec:model-generality}

To assess architectural generality, we apply the same training recipe on \textbf{Llama‑3.1‑8B‑Instruct}.  
Table~\ref{tab:llama-general} shows that the RL model again reduces \textsc{MSE} relative to its SFT counterpart, mirroring the improvements observed for the Qwen backbone.  
These findings confirm that the \textsc{ReviewRL} training recipe generalizes across model families.

\end{document}